%% file: acl_latex.tex
\pdfoutput=1

\documentclass[11pt]{article}

\usepackage[final]{acl}
\usepackage{hyperref}
\usepackage{times}
\usepackage{latexsym}
\usepackage{multirow}
\usepackage{booktabs}
\usepackage{microtype}
\usepackage{algorithm}
\usepackage{algorithmic}
\usepackage{multirow}
\usepackage{latexsym}
\usepackage{booktabs}
\usepackage{courier}
\usepackage{amsmath}
\usepackage{booktabs} 
\usepackage{siunitx} 
\usepackage{subfig}
\usepackage{placeins}
\usepackage{tikz}
\usepackage{hyperref}
\usepackage{multicol}

\usepackage[T1]{fontenc}

\usepackage[utf8]{inputenc}

\usepackage{microtype}

\usepackage{inconsolata}
\usepackage{soul}

\title{IITK at SemEval-2024 Task 1: Contrastive Learning and Autoencoders for Semantic Textual Relatedness in Multilingual Texts}

\author{Udvas Basak\thanks{\ \ Equal Contributions} \qquad Rajarshi Dutta\footnotemark[1] \qquad Shivam Pandey\footnotemark[1] \qquad Ashutosh Modi  \\
Indian Institute of Technology Kanpur (IIT Kanpur)\\
\texttt{\{udvasb20, rajarshi20, shivamp20\}@iitk.ac.in} \\ \texttt{ashutoshm@cse.iitk.ac.in} 
}

\begin{document}

\maketitle

\input{sections/abstract}
\input{sections/introduction}
\input{sections/background}
\input{sections/dataset}
\input{sections/system}
\input{sections/experiments}
\input{sections/results}
\input{sections/conclusion}

\bibliography{custom}

\appendix
\input{sections/appendix}

\end{document}

%% file: sections/abstract.tex
\begin{abstract}
       This paper describes our system developed for the SemEval-2024 Task 1: Semantic Textual Relatedness. The challenge is focused on automatically detecting the degree of relatedness between pairs of sentences for 14 languages including both high and low-resource Asian and African languages. Our team participated in two subtasks consisting of Track A: supervised and Track B: unsupervised. This paper focuses on a BERT-based contrastive learning and similarity metric based approach primarily for the supervised track while exploring autoencoders for the unsupervised track. It also aims on the creation of a bigram relatedness corpus using negative sampling strategy, thereby producing refined word embeddings. 

\end{abstract}

%% file: sections/introduction.tex
\section{Introduction} \label{sec:intro}

The semantic relatedness between texts in a language is fundamental to understanding meaning \cite{halliday&hasan}. Automatically detecting relatedness plays an essential role in evaluating sentence representations, question answering, and summarization \cite{abdalla-etal-2023-makes}. The fundamental difference between semantic similarity and relatedness is that semantic similarity only considers paraphrase or entailment relationships. In contrast, relatedness accounts for all commonalities between two sentences, e.g., topical, temporal, thematic, contextual, syntactic, etc. \cite{abdalla-etal-2023-makes}. As highlighted in Table \ref{table:sentence_egs}, Sentences 1 and 2 are semantically similar, but sentences 2 and 3 would have low semantic similarity but high semantic relatedness.

\begin{table}[t]
\centering
\resizebox{\columnwidth}{!}{
\begin{tabular}{@{}lc@{}}
\toprule
\textbf{\#} & \textbf{Sentence}\\
\midrule
1 & The mouse was chased by the cat in the yard.\\
2 & The cat chased the mouse around the garden.\\
3 & The dog barked loudly as the mouse scurried away.\\
\bottomrule
\end{tabular}%
}
\caption{Difference between Similarity and Relatedness}
\label{table:sentence_egs}
\vspace{-4mm}
\end{table}

In Track A (Task 1) of the Semantic Textual Relatedness (STR) task \cite{semrel2024task}, we are expected to calculate the degree of semantic relatedness between pairs of sentences in 14 different languages covering both African and Asian languages. Each pair of sentences is assigned a relatedness score in the range of 0 and 1. The major challenge lies in the efficient development of a metric to facilitate the calculation of the relatedness score between the sentence pairs and harnessing the structure of multiple languages to create an efficient model \cite{semrel2024task}. Our system is based on a contrastive learning approach, utilizing a composite lexical similarity-based measure for relatedness score calculation in the supervised task. Additionally, it involved the use of transformer autoencoders for the unsupervised task. We employed Distill-RoBERTa \cite{sanh2020distilbert} as the model for this purpose. Several other strategies were also tested within this framework, such as employing a Siamese architecture and retraining BERT with vocabulary expansion to incorporate tokens from additional low-resource languages. For the unsupervised task, the base model used to construct the denoising autoencoder was BERT-uncased \cite{devlin2019bert}. The major challenge in this task was the devising and implementing data pre-processing schemes for diverse languages and various training methodologies. A number of experiments were conducted to come up with a unified metric for semantic relatedness calculations, which resulted in relatively better performances in various low-resource languages.\footnote{The code can be found at \href{https://github.com/Exploration-Lab/IITK-SemEval-2024-Task-1}{https://github.com/Exploration-Lab/IITK-SemEval-2024-Task-1-Semantic-Relatedness}}


%% file: sections/background.tex
\section{Background} \label{sec:background}

There have been several attempts to define and distinguish semantic relatedness from semantic similarity. The basic metric used in these experiments is \href{https://en.wikipedia.org/wiki/Spearman%27s_rank_correlation_coefficient}{Spearman Rank Correlation}. The correlation coefficient is calculated between the correctly annotated scores for the set of pairs of sentences and the scores returned by the models. Essentially, this removes the absolute values of the scores and focuses on the relative values and, hence, the relative relatedness between the pairs of sentences.

Initial experiments explored frequency measures such as lexical overlap \cite{shirude-etal-2021-iitk}, related words, and related subjects and objects, leading to high Spearman correlations of $0.82$ and $0.83$ for BERTbase (mean) and RoBERTa-base (mean) on the CompLex dataset \cite{shardlow-etal-2020-complex}. Despite marginal improvements over a lexical overlap baseline, unsupervised models offer limited enhancement. 

Normalized Google Distance(NGD) has been used as a novel metric for measuring semantic relatedness between words or concepts \cite{8760964}, utilizing Google search result counts to quantify relatedness. NGD \cite{ngd_original} normalized counts considering the overall corpus size and the co-occurrence of terms in web pages. 

Various approaches have been proposed for the Arabic language \cite{arabicbert} like automatic machine translation to translate English Semantic Textual Similarity (STS) data into Arabic, interleaving English STS data with Arabic BERT models, and employing knowledge distillation-based models to fine-tune them using a translated dataset. Multilingual knowledge distillation \cite{multilingual} techniques have been proposed where a student model, $\hat{M}$, learns from a teacher model, $M$, on source language sentences and their translations by minimizing the mean-squared loss function. Focusing on low-resource Indian languages, a range of SBERT models has been introduced for ten popular Indian languages. IndicSBERT \cite{Indicbert:01} utilized a two-step training method, fine-tuning models using the NLI dataset followed by Semantic Textual Similarity benchmarking (STSb), resulting in substantial improvements in embedding similarity scores and cross-lingual performance.


%% file: sections/dataset.tex
\section{Dataset Description}

The dataset \cite{semrel2024dataset} consists of a total of 14 languages, namely Afrikaans, Algerian Arabic, Amharic, English, Hausa, Indonesian, Hindi, Kinyarwanda, Marathi, Modern Standard Arabic, Moroccan Arabic, Punjabi, Spanish, and Telugu. Every language consists of pairs of sentences with scores representing the degree of semantic textual relatedness between 0 and 1. The scores have been assigned to sentence pairs through a comparative annotation process \cite{semrel2024dataset}. 

At the preliminary level, dataset length is the only non-semantic variable in these datasets. To assess semantic relatedness, it is crucial to mitigate these biases. Correlation coefficients between sentence lengths and scores fall in the range $-0.13 < \rho < 0.15$; hence, there is no discernible correlation between sentence lengths and scores, suggesting a well-distributed dataset suitable for training.

%% file: sections/system.tex
\section{System Overview}\label{sec:system}

\begin{figure}[t]
    \centering
    \includegraphics[width=0.20\textwidth]{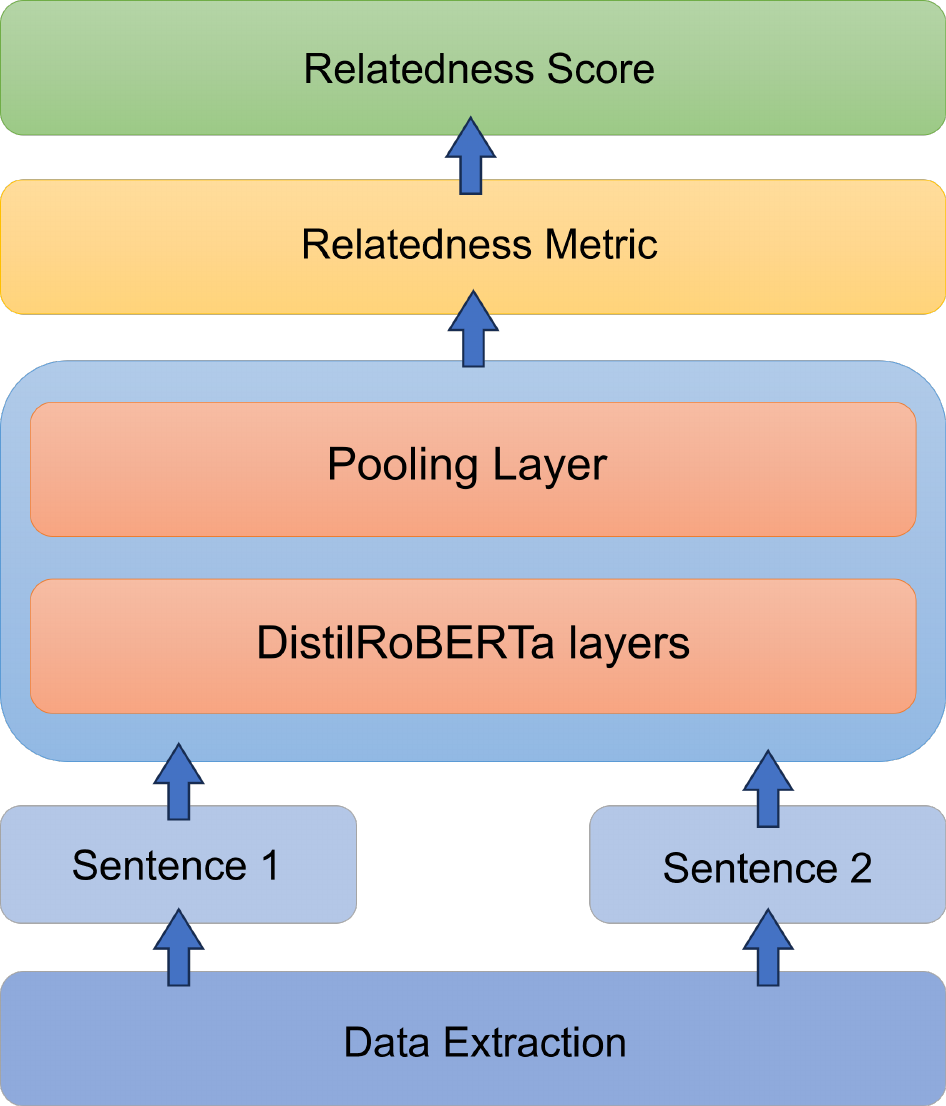}
    \caption{SIMCSE based approach for Track A}
    \label{fig:simcse_arch}
    \vspace{-4mm}
\end{figure}

Our baseline system of scoring a pair of sentences uses \href{https://en.wikipedia.org/wiki/Jaccard_index}{Jaccard Similarity}, a lexical metric that calculates the number of token intersections over total tokens in both sentences. Our approach involves using Contrastive learning for the supervised part and auto-encoders for the unsupervised part. All these methods are discussed below, and our model supervised architecture is shown in Figure \ref{fig:simcse_arch}. 

\subsection{SimCSE}\label{subsec:SIMCSE}
SimCSE \cite{gao2022simcse}, or Simple Contrastive Learning, is helpful in supervised and unsupervised learning, particularly in information retrieval, text clustering, and semantic tasks. This approach primarily uses Natural Language Inference (NLI) to create positive and negative sentence samples. It works by inducing slight variation in its representations through dropouts. The following step lies in aligning related sentences close in the embedding space and maximizing distances to unrelated sentences in each batch of data. For a supervised setting, it classifies positive samples as entailment pairs, while negative samples are derived from contradiction pairs. The training proceeds via minimizing the loss function: $-\log\left(\frac{e^{\frac{sim(h, h^{+})}{\tau}}}{e^{\frac{sim(h, h^{+})}{\tau}} + e^{\frac{sim(h, h^{-})}{\tau}}}\right)$, where, $h$ represents the current sentence and $h^{+}$ and $h^{-}$ denotes the positive and negative samples respectively with $\tau$ being the temperature hyperparameter which controls the sensitivity and learning dynamics. The $\tau$ is mainly adjusted based on validation set performance. 



\subsection{TSDAE}\label{subsec:TSDAE}

TSDAE \cite{wang2021tsdae} or Transformer Denoising through Auto Encoders is an elegant approach aimed at improving the quality of sentence embeddings through self-supervised learning. This method mainly uses sentence embeddings without labels, relying solely on the data structure to initiate learning. The first step lies in corrupting the existing tokens via methods like deleting random tokens, masking tokens, etc., and passing these modified representations through an encoder layer. The encoder layer outputs a dense latent representation, capturing the essence of the data in high-dimensional space. The encoded representations are then passed to a decoder, which attempts to reconstruct the original, uncorrupted sentences from the encoded representations. The decoder is typically another transformer model that has been trained to generate text based on the encoded embeddings. The main objective lies in minimizing the distance between the corrupted and reconstructed sentence representations through cross-entropy loss. 

\subsection{Training Scheme}

For the supervised track, we used Distil-RoBERTa \cite{sanh2020distilbert} as the base model, which produced a vector representation for every word in the input sentence, resulting in a matrix of token embeddings. The embeddings were then fed to a pooling layer for the production of sentence-level embeddings, which finally proved to be useful for semantics-relatedness tasks. We used mean pooling because there were no dedicated [CLS] token representations for sequence classification tasks. As for the metrics, our approach was involved in designing a custom relatedness metric by combining standard distance-based metrics like cosine similarity, Mahalanobis distance, Euclidean and Manhattan distances, and lexical overlap-based metrics like Jaccard and Dice coefficients. For each pair of sentence embeddings in the dataset, we calculated these metrics. Not only did we calculate these metrics using the original embeddings, but we also calculated them after transforming the embeddings by raising them to higher powers (e.g., squaring them). These calculated metrics were then collected into a dataset, with each column named according to the metric and the power applied to the embeddings. For example, the column ``Cosine Distance: 2'' depicted the cosine distances between pairs of sentence embeddings after both embeddings in each pair have been squared. The dataset, therefore, finally had rows where each row was a 42-element vector. This vector encompassed the calculated metrics across different powers for the sentence embeddings. These enhanced sentence embeddings, with metrics covering higher orders, were then used to train the RoBERTa model. The goal was to produce scores that indicate how related different sentences are across various languages. The libraries used are in Table \ref{table:libraries}.

%% file: sections/experiments.tex
\section{Experiments} \label{sec:experiments}
\subsection{Supervised Task}

\textbf{Static Approaches:} The baseline models, \href{https://en.wikipedia.org/wiki/Jaccard_index}{Jaccard Coefficient}, \href{https://en.wikipedia.org/wiki/S%C3%B8rensen%E2%80%93Dice_coefficient}{Dice Coefficient}, and similar coefficients after removing stopwords were calculated to arrive at reliable baseline metrics to build upon.


\noindent\textbf{Multilingual BERT:} Since the best-performing model for English involved BERT, an attempt was made to train multilingual BERT: mBERT \cite{pires2019multilingual}, by extending the vocabulary to allocate the tokens of various low-resource languages like Amharic, Hausa, Algerian Arabic, Afrikaans, Indonesian etc. The approach included generation of the vocabulary of each of the languages from the training data and then calling the pre-trained mBERT model and tokenizer. The trained tokenizer was extended to include the new tokens generated from the vocab of the corresponding low
resource languages. A trainable feed-forward network was added with the corresponding dropout. The loss metric used was mean squared error loss on both the training validation data and the Spearman rank correlation were calculated at the end of each validation epoch. Finally, finetuning multilingual BERT yielded considerably good results and this avenue was found suitable for exploration, especially for the cross-lingual task.

The relatedness metric was approximated using a trainable feed forward layer by experimenting with the number of hidden layers and activations. It was observed that having 3 hidden layers resulted in fairly good relatedness scores between pairs of sentences in almost all the languages. It was observed that GeLU performed better than ReLU activation primarily due to its steep curve around 0 which helps to model complex functions better. The combination of learning rate and weight decay also resulted in a stable training curve, avoiding suboptimal loss convergence. Thus these specific hyperparameters were optimal for mBERT retraining in terms of resource constraints and performance. The corresponding hyperparameters of the best performing model is presented in App. Table \ref{table:mBERT_hyperparams}.


\noindent\textbf{Contrastive Learning:} The details of the system are described in \S\ref{subsec:SIMCSE}. Experiments were run on the number of epochs during training. 

\begin{figure}[t]
    \centering
    \includegraphics[width=0.30\textwidth]{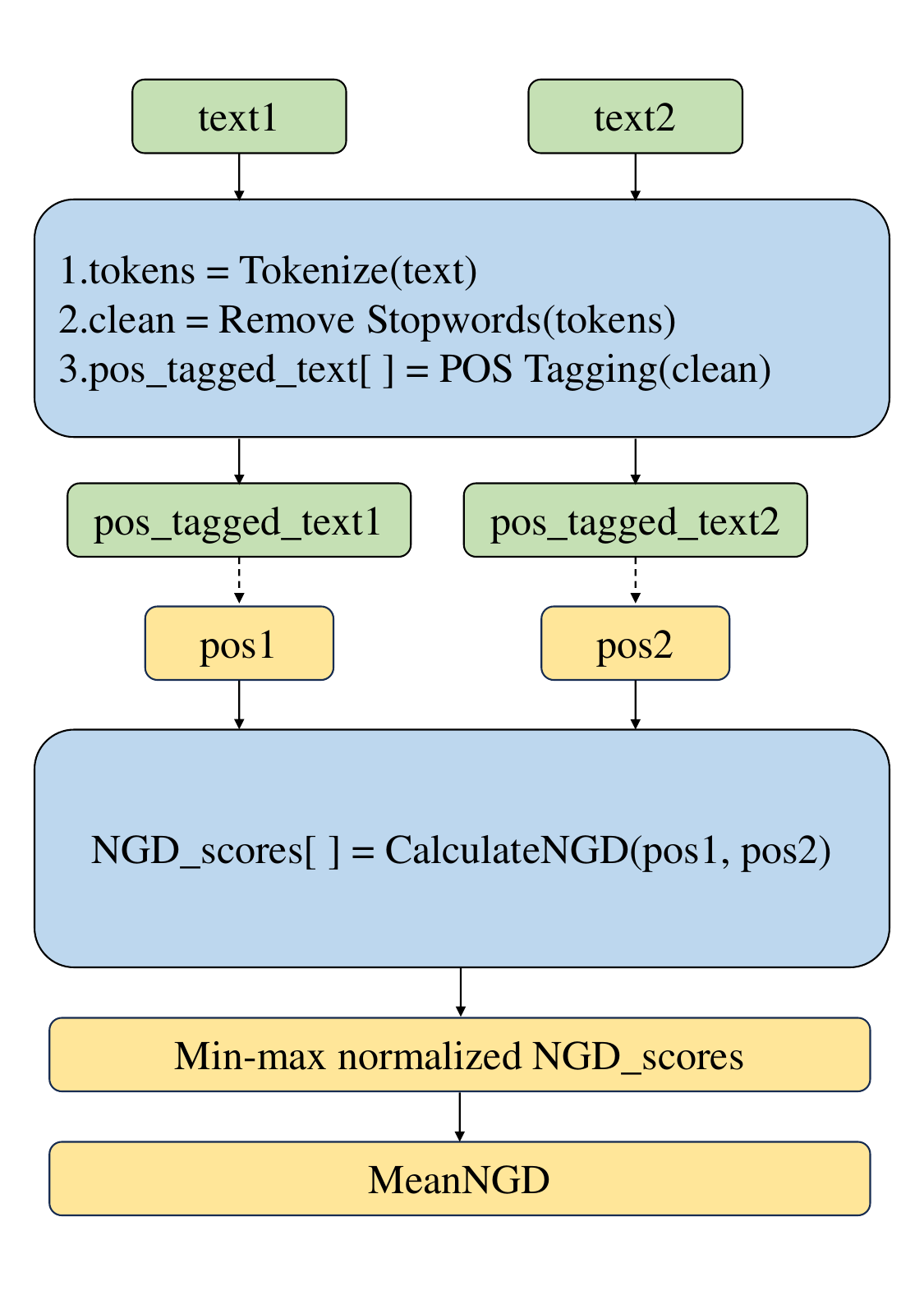}
    \caption{NGD Calculation flowchart}
    \label{fig:ngd_flowchart}
    \vspace{-5mm}
\end{figure}
\begin{figure}[t]
    \centering
    \includegraphics[width=0.40\textwidth]{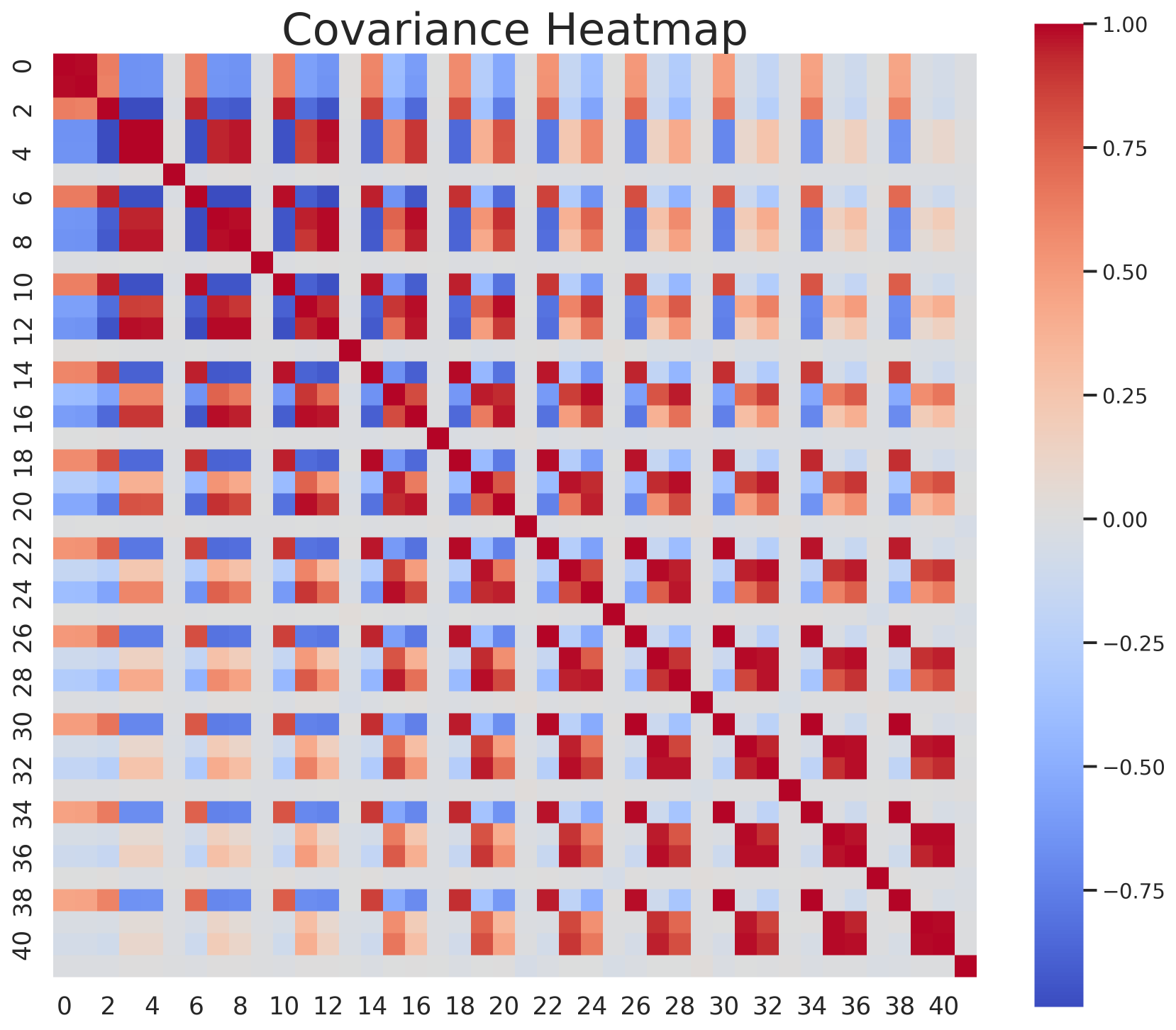}
    \caption{Covariance Matrix between all 42 metrics}
    \label{fig:metric_cov}
    \vspace{-4mm}
\end{figure}

\noindent\textbf{Combined Similarity Metric:} Normalized Google Distance \cite{ngd_original} calculates a relatedness metric for two input sentences. It was proposed as a strong metric for relatedness by \citet{8760964}. It starts by tokenizing and removing stop words from both sentences, followed by part-of-speech tagging. Then, it calculates NGD values for pairs of words with the same part of speech in both sentences. The NGD scores are normalized and averaged to compute the overall NGD score, representing the degree of relatedness between the two sentences. The flowchart for the process is shown in Figure \ref{fig:ngd_flowchart}. 
Cosine similarity is the standard metric used to find similarity between two sentence embeddings, which gives a 0.81-0.82 baseline score for this problem.However, similar or better results can be seen when other distance metrics like Mahalanobis Distance(0.82) and Euclidean Distance(0.83) are observed between the embeddings. Further, augmenting this with more direct relatedness metrics like NGD is promising for better results. A simple supervised deterministic regression model can be implemented to combine these metrics. Furthermore, to explore the importance of each of these metrics, a simple covariance matrix (Figure \ref{fig:metric_cov}) can show how the vector metrics on higher element-wise-powered vectors hold information not caught directly at the lower powers of the vectors.

\normalsize

To implement this supervised regression model, a simple 3-layered feed-forward neural network (with neurons \textbf{[25]+[50]+[25]}) is trained with. The layers are chosen to construct a lightweight network. Each data feature $\mathbf{x}$ was composed as:

\small{
\begin{align*}
\mathbf{x_i} = \{ & S(v_{i,1}, v_{i,2}), S(v^2_{i,1}, v^2_{i,2}), \ldots, \\
& S(v^{10}_{i,1}, v^{10}_{i,2}), J(v_{i,1}, v_{i,2}), D(v_{i,1}, v_{i,2}) \} \\
\text{where} \quad & v^i = (v_1^i, v_2^i, \hdots v_n^i)\\
& S(a,b) = \{C(a,b), E(a,b), M_{1}(a,b), M_{2}(a,b)\}\\
& C(a,b) = \text{Cosine Similarity between } a \text{ and } b\\
& E(a,b) = \text{Euclidean Distance between } a \text{ and } b\\
& M_{1}(a,b) = \text{Manhattan Distance between } a \text{ and } b\\
& M_{2}(a,b) = \text{Mahalanobis Distance between } a \text{ and } b\\
& J(a, b) = \text{Jaccard similarity between } a \text{ and } b\\
& D(a, b) = \text{Dice similarity between } a \text{ and } b
\end{align*}}
\normalsize
Even though this metric did not show promise in English, this was helpful in some low-resource languages, and hence was part of our system design for some languages.

\subsection{Unsupervised Task}

\begin{figure}
    \centering
    \includegraphics[width=0.30\textwidth]{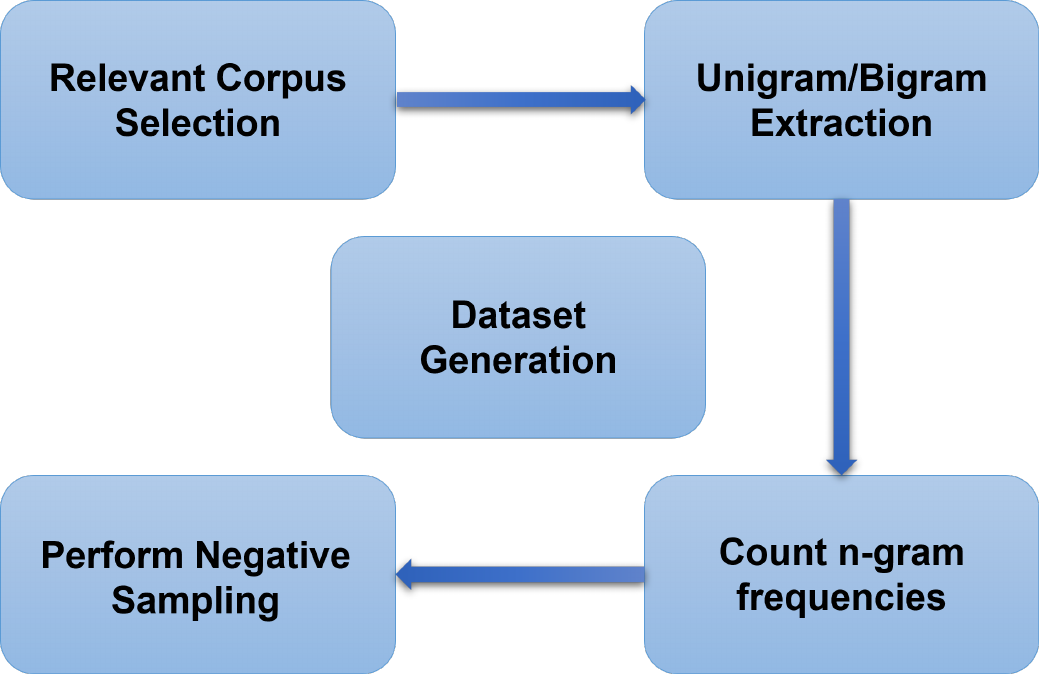}
    \caption{Bigram Corpus Creation Flowchart}
    \label{fig:corpus_flowchart} 
    \vspace{-4mm}
\end{figure}

\textbf{Bigram Corpus Creation and Training Process}: We developed a pipeline to generate a bigram dataset from any language corpus. A three-part tuple was created for every bigram found to note how often it appeared within the same sentence, paragraph, and entire document. This process aimed to quantify the connections between words by tracking their repeated sentence occurrences. Our main objective was to use these co-occurrence frequencies to produce word embeddings. We planned to enhance our method by applying hierarchical clustering, which helps identify word similarities and relationships. Moreover, we decided to use a 1:1 \textbf{negative sampling strategy} to refine the embeddings further. These embeddings were intended for computing relatedness scores, leveraging bigrams derived from sentence pairs and their lexical overlaps. Figure \ref{fig:corpus_flowchart} presents a diagram illustrating this process.

\vspace{0.5em}

\noindent\textbf{TSDAE:} A pipeline was developed to implement TSDAE (refer \S \ref{subsec:TSDAE}) on the languages. The number of epochs was changed and experimented on for various languages. Overall, around 20-25 epochs resulted in good results. No weight decay was implemented, and a learning rate of 3e-5 was used. 

%% file: sections/results.tex
\section{Results} \label{sec:results}

The results for the supervised and unsupervised are mentioned in Table \ref{table:metrics} and Table \ref{table:metrics_contrast}. The leaderboard highlights that the chosen contrastive learning approach did not perform well, especially for some languages where it fell significantly below the baseline scores provided by their system, such as Hausa, Moroccan Arabic, Telugu, etc. Some possible shortcomings of this approach might be that the negative samples were not distinguishable enough from the positive samples, which might also be attributed to the poor performance of the transformer models on languages, especially with complex lexical structures. The other issue might be the traditional loss function, which might not be good enough to capture the degree of semantic relationships between sentences. For the unsupervised track, our approach is performing reasonably well for most languages, which is indicated by the correlation score being more significant than the baseline provided. 

\section{Error Analysis} \label{sec:error_analysis}

After the evaluation phase, we were provided with the labels for the evaluation data. On experimenting with the semantic relatedness scores for some languages, mainly \textbf{Hausa} and \textbf{Kinyarwanda}, we found out that our system was not performing well enough on these languages even after subsequent training and hyperparameter optimizations. The issue would be primarily attributed to the complex lexical variations and grammar rules of these languages. As for the unsupervised track, generating a bigram corpus for the case of \textbf{Amharic} seemed difficult due to its language structure. 

\begin{table}[t]
\centering
\tiny
\resizebox{\columnwidth}{!}{
\begin{tabular}{@{}lccc@{}}
\toprule
\textbf{Language} & \textbf{Rank} & \textbf{Score} & \textbf{Baseline Score} \\
\midrule
Amharic & 17 & 0.55 & 0.85 \\
Hausa & 21 & 0.22 & 0.69 \\
Kinyarwanda & 21 & 0.14 & 0.72 \\
Moroccan Arabic & 22 & 0.36 & 0.77 \\
Spanish  & 23 & 0.59 & 0.7 \\
Algerian Arabic & 23 & 0.34 & 0.6 \\
Marathi & 24 & 0.67 & 0.88 \\
Telugu & 25 & 0.28 & 0.82 \\
English & 31 & 0.81 & 0.83 \\
\bottomrule
\end{tabular}%
}
\caption{Evaluation Phase Results in Codalab Leaderboard for Track A}
\label{table:metrics}
\vspace{-3mm}
\end{table}

\begin{table}[t]
\centering
\tiny
\resizebox{\columnwidth}{!}{
\begin{tabular}{@{}lcccc@{}}
\toprule
\textbf{Language} & \textbf{Rank} & \textbf{Score} & \textbf{Baseline Score} \\
\midrule
Algerian Arabic & 2 & 0.49 & 0.43 \\
English & 4 & 0.81 & 0.68 \\
Amharic & 6 & 0.07 & 0.72 \\
Hausa & 6 & 0.38 & 0.16 \\
Moroccan Arabic & 6 & 0.36 & 0.27 \\
Spanish  & 9 & 0.59 & 0.69 \\
\bottomrule
\end{tabular}%
}
\caption{Evaluation Phase Results in Codalab Leaderboard for Track B}
\label{table:metrics_contrast}
\vspace{-5mm}
\end{table}

%% file: sections/conclusion.tex
\section{Conclusion} \label{sec:conclusion}

By utilizing various approaches like contrastive learning, autoencoders, a custom relatedness metric incorporating all of the available lexical similarity metrics, we have developed a  system capable of evaluating the degree of semantic relatedness between pairs of sentences in diverse high and low resource languages. In future, we will study the properties of each low resource language to find out where the models are performing poorly than relying too much on pre-trained models. This would give much clearer insights into semantics of each language thus improving the overall efficiency and performance of our system.

%% file: sections/appendix.tex
\section{Appendix} \label{sec:appendix}

\begin{table}[h]
\centering
\resizebox{0.7\columnwidth}{!}{
\begin{tabular}{@{}lc@{}}
\toprule
\textbf{Hyperparameters} & \textbf{Values}\\
\midrule
Learning Rate & 2e-5\\
Dropout & 0.1\\
Weight Decay & 0.01\\
Number of Linear Layers & 3\\
Activation & GELU\\
Max Length & 512\\
\bottomrule
\end{tabular}%
}
\caption{Hyperparameters for mBERT retraining}
\label{table:mBERT_hyperparams}
\end{table}

\begin{table}[h]
\centering
\resizebox{0.7\columnwidth}{!}{
\begin{tabular}{@{}lc@{}}
\toprule
\textbf{Libraries} & \textbf{Version}\\
\midrule
numpy & 1.25.2\\
PyTorch & 2.0.1+cu117\\
transformers & 4.36.2\\
sentence\_transformers & 2.2.2\\
scikit-learn & 1.3.2\\
pandas & 2.1.4\\
\bottomrule
\end{tabular}%
}
\caption{Libraries used in our system}
\label{table:libraries}
\end{table}